%% file: main.tex
\documentclass[10pt,twocolumn,letterpaper]{article}

\usepackage[pagenumbers]{wacv} % To force page numbers, e.g. for an arXiv version

\usepackage{xcolor}
\usepackage{graphicx}
\usepackage{multirow}
\usepackage{amsmath}
\usepackage{xspace}
\newcommand{\name}[0]{{\sc FlashSwin}\xspace}
\input{preamble}

\definecolor{wacvblue}{rgb}{0.21,0.49,0.74}
\usepackage[pagebackref,breaklinks,colorlinks,allcolors=wacvblue]{hyperref}

\title{\name: Unlocking Large Windows and Dense Tokens in Swin Vision Transformers with Memory Efficient Attention}

\author{\textbf{Tushar Kataria}$^{1,2}$
\quad
\textbf{Gerald Sabin}$^{3}$
\quad
\textbf{Ponnuswamy Sadayappan}$^{2}$
\quad
\textbf{Shireen Y. Elhabian}$^{1,2}$\\
$^{1}$ Scientific Computing and Imaging Institute, University of Utah, Salt Lake City, UT, USA \\
$^{2}$ Kahlert School of Computing, University of Utah, Salt Lake City, UT, USA  \\
$^{3}$ RNET Technologies, Dayton, OH, USA\\
{\tt\small tushar.kataria@utah.edu,gsabin@rnet-tech.com, saday@cs.utah.edu, shireen@sci.utah.edu} 
}

\begin{document}
\maketitle
\input{sec/0_abstract}    
\input{sec/1_intro}
\input{sec/2_related_works}
\input{sec/3_methods}
\input{sec/4_results}
\input{sec/5_conclusion}
{
    \small
    \bibliographystyle{ieeenat_fullname}
    \bibliography{main}
}
\clearpage
\input{sec/6_supp}

\end{document}

%% file: preamble.tex
\definecolor{gmsblue}{rgb}{0.10,0.30,0.65}
\usepackage[
  textsize=footnotesize,
  backgroundcolor=gmsblue!8,
  bordercolor=gmsblue!55,
  linecolor=gmsblue!55
]{todonotes}

%% file: sec/0_abstract.tex
\begin{abstract}
High-resolution vision backbones have long been forced to trade away local token density to afford larger receptive fields. Hierarchical Swin transformers impose this compromise because standard windowed attention materializes an $M^2\times M^2$ score matrix per window, incurring $O(M^4)$ memory as windows or token grids grow. Furthermore, Swin adds a learned relative-position bias elementwise to attention scores, requiring full materialization of the score matrix and its gradient. This keeps Swin and SwinV2 trapped in a small-window($M=8,16$), coarse-token regime with patch size $4\times4$ ($p=4$), limiting performance for fine-grained tasks.
We introduce \name{}, which replaces standard windowed attention with a FlashAttention implementation that computes exact softmax attention without materializing the score matrix, reducing per-window memory from $O(M^4)$ to $O(M^2)$. This enables higher token density and larger receptive fields without inflating memory overhead. Peak training memory becomes virtually flat in window size: at a $32\times32$ window, \name-T requires only $12.4$\,GB—unchanged from $8\times8$—compared to $70/90$\,GB for SwinV2/V1-T. However, applying FlashAttention directly to Swin creates a trade-off: bypassing the score matrix precludes Swin's additive relative-position bias, forfeiting spatial information in exchange for memory efficiency. \name{} restores position information as window-local learnable 2D RoPE, making large windows and dense token grids both affordable and accurate. At matched scale, \name-T outperforms Swin variants while improving out-of-domain generalization. 
With dense tokens and wide windows ($p=2,M=32$), the same Tiny model reaches $84.1\%$ ImageNet-1K, $44.1$ COCO box AP, and $47.28$ ADE20K mIoU---gains of $+1.3$, $+5.1$, and $+1.82$ over SwinV2-T at $M=16$, respectively. At fixed $M=32$, halving the patch size yields roughly $3\times$ larger gains in boundary quality than in mIoU.
By removing memory bottlenecks, \name{} unlocks a new scaling axis for swin transformers: combining dense tokens with large receptive fields.
\end{abstract}

%% file: sec/1_intro.tex
\section{Introduction}
\label{sec:intro}
Fine-grained spatial detail is essential to a wide range of vision tasks, yet vision-transformer backbones~\cite{dosovitskiy2020image} are designed for and trained at comparatively low resolution. Dense prediction tasks such as semantic segmentation and object detection depend on this detail to localize boundaries and resolve small structures~\cite{chen2022vision,zhang2022segvit,zhang2024segvit,liu2021swin}. The gap is widest in domains whose native inputs dwarf the $224^2$ or $256^2$ images used to pretrain modern backbones: gigapixel histopathology slides, volumetric medical scans, aerial imagery, and document scans~\cite{neidlinger2025benchmarking,vorontsov2024foundation,xu2025generalizable,simeoni2025dinov3,da2023vision}. In each case the standard remedy is to discard resolution at the input itself. Patch embedding collapses a $16\times16$ region into a single token in ViT, a $4\times4$ region in Swin~\cite{liu2021swin}, and $16^3$ or $4^3$ voxel blocks in 3D equivalents~\cite{xu2025generalizable,bin2025efficientmorph,chen2022transmorph}. The size of these patches is dictated by the memory and compute cost of the backbone rather than by the granularity the task requires. Hierarchical windowed transformers such as Swin~\cite{liu2021swin} were introduced precisely to relieve this pressure, restricting attention to local windows so that cost grows linearly, rather than quadratically, in the number of tokens.

That linear scaling, however, has not translated into high-resolution training. Three coupled choices govern how much spatial information a windowed transformer actually captures: the window size $M$, the patch size of the stem, and the input resolution. The latter two jointly fix the token grid---a finer patch or a larger input yields denser tokens that retain local structure and reduce the need for auxiliary refinement modules---while $M$ fixes the region over which those tokens interact directly, before information must propagate through the hierarchy. Token density and window size act along complementary axes, and together they set the balance between local detail and contextual range available to the model~\cite{yang2021focal,dong2022cswin}. In practice all three are kept small. Swin~\cite{liu2021swin} and SwinV2~\cite{liu2022swin} train with narrow windows ($M = 7$-$8$ at pretraining resolution), a coarse patch-$4$ stem, and $224^2$-$256^2$ inputs, fine-tuning to higher resolution for only $30$ epochs against $300$ for pretraining, and at considerable cost. Crucially, these settings are not chosen for accuracy; they are forced by \emph{memory}. Windowed attention materializes an $M^2 \times M^2$ score matrix per window, an $O(M^4)$ footprint; summed over the $HW/M^2$ windows of a feature map this becomes $O(HW M^2)$---a factor of $M^2$ more entries than tokens, growing with the window \emph{and} the token count, and therefore quadratically in the input side length. High-resolution training thus reaches a memory ceiling of the available hardware well before its accuracy gains are exhausted. The result is that dense token grids, wide windows, and native high-resolution training remain largely unexplored. % because that corner of the design space does not \emph{fit}, not because it does not \emph{work} (\cref{fig:token_arch_memory}C).

FlashAttention~\cite{dao2022flashattention} removes exactly this bottleneck. %Computing attention in tiles, 
It computes the attention output incrementally using tiled online softmax without fully materializing the score matrix in memory, cutting the per-window footprint from $O(M^4)$ to $O(M^2)$. The kernel computes the same softmax attention, without approximation.
Window size in particular stops driving activation memory, though token count still does. Swin, however, cannot adopt the FlashAttention kernel as a drop-in replacement. Its relative-position bias is added elementwise to the $M^2 \times M^2$ score matrix, the exact tensor the fused kernel avoids forming to save memory. 
Retaining the bias forfeits memory savings, while dropping it degrades accuracy by removing intra-window positional awareness (\cref{tab:abl-comp}).
Nor does enlarging the bias help: Swin's table covers only the $(2M-1)^2$ offsets seen in training, and although SwinV2's continuous reformulation does generate unseen offsets, it remains a learned term added to the $M^2 \times M^2$ scores-and a learned term there needs its gradient there. The obstacle is additivity, not coverage.  Recovering Swin's accuracy under exact, memory-efficient attention therefore requires moving position out of the score matrix. We introduce \name{}, which leaves the hierarchy intact but replaces the two components that assume a materialized matrix or a coarse grid: the position bias becomes a 2D RoPE applied to queries and keys before the FlashAttention kernel, continuous in the coordinate difference so it evaluates at any window size, with learned frequencies; the disjoint patch embedding becomes an overlapping one, preserving detail across patch boundaries at high token density. The savings are large (\cref{fig:token_arch_memory}C): at a $32\times32$ window, \name{}-T trains in $12.4$GB against $70.4$GB for SwinV2-T, and it trains grids $4\times$ denser than the standard patch-$4$ setting, which at this window exhaust memory for both Swin variants.

At matched compute ($5.9$ GFLOPs), \name{}-T achieves $82.3\%$ ImageNet-1K top-1 accuracy, outperforming Swin baselines with gains transferring consistently to ADE20K and COCO. Scaling to dense tokens and wider windows ($p=2, M=32$) elevates the same $28$M backbone to $84.1\%$. However, clean classification understates the broader impact: \name{}-T improves robustness to adversarial, sketch, and corrupted inputs, while its relative gains on ADE20K boundary metrics are approximately $3\times$ larger than on mIoU. These results suggest that spatial resolution has a particularly strong effect on robustness and fine localization. This denser regime increases compute to $37.4$ GFLOPs, but remains trainable at batch sizes where standard windowed attention exceeds memory limits and requires activation checkpointing, which saves memory at the cost of computation~\cite{chen2016training}. Our contributions are:
\begin{itemize}
\item \textbf{\name{}: Exact fused attention for the hierarchical networks.} Window-local 2D RoPE and overlapping tokens make standard FlashAttention usable inside Swin. \name-T achieves $82.3\%$ ImageNet-1K top-1 at matched compute ($5.9$ GFLOPs, vs.\ $81.9\%$ SwinV2-T), with better out of distribution generalization.

\item \textbf{Spatial resolution governs boundary fidelity and robustness.} Our results show that contextual range dictates region classification, while token density dictates boundary localization—yielding $3\times$ larger relative boundary gains on ADE20K than mIoU. At fixed $28$M capacity, expanding spatial resolution boosts out-of-distribution accuracy, a pattern that scales consistently to large models ($\approx 197$M parameters). \href{https://github.com/tushaarkataria/FlashSwin}{https://github.com/tushaarkataria/FlashSwin.}
\end{itemize}

%% file: sec/2_related_works.tex
\section{Related Works}
\label{sec:related_works}

\textbf{Hierarchical and windowed vision transformers.} Vision Transformers apply self-attention over image patches, but quadratic complexity makes high-resolution processing expensive~\cite{dosovitskiy2020image,liu2021swin,dong2022cswin,yang2021focal,wang2021pyramid}. Hierarchical architectures address this by progressively reducing spatial resolution and restricting attention to local regions. Swin~\cite{liu2021swin} uses shifted non-overlapping windows to achieve linear complexity, while SwinV2~\cite{liu2022swin} improves scaling through normalization, attention changes, and continuous relative-position biases. Others redesign the attention neighborhood: Focal Transformer~\cite{yang2021focal} combines local interactions with coarse global context, CSWin~\cite{dong2022cswin} uses cross-shaped windows, and NAT~\cite{hassani2023neighborhood} uses sliding neighborhoods. A complementary line reduces token counts through token merging~\cite{bolya2022token}, simplified hierarchies~\cite{ryali2023hiera}, or post hoc feature pyramids~\cite{li2022exploring}. In contrast, \name{} retains Swin's hierarchy and attention pattern, targeting memory costs \emph{within} each window so window size and token density can expand rather than economize—preserving fine-grained spatial details without prohibitive memory overhead.

\textbf{Memory-efficient attention and positional encodings.}
FlashAttention~\cite{dao2022flashattention,dao2023flashattention2} preserves exact softmax attention through IO-aware tiling without materializing the attention matrix. Integrating fused attention into windowed vision transformers, however, is complicated by Swin's additive score-level positional bias. One approach is to modify the fused kernel itself: FlashBias~\cite{wu2026flashbias} incorporates additive biases during tiled attention, although its default formulation is not end-to-end trainable and requires a low-rank approximation for training. An alternative is to move positional modeling outside the kernel by encoding position directly in queries and keys.
Following this second direction, Heo et al.~\cite{heo2024rotary} extend RoPE to vision using 2D positional coordinates and learnable mixed frequencies. \name{} builds on this formulation but uses a single frequency set shared across both spatial axes and attention heads, with positions indexed locally within each window. This yields a window-size-independent positional representation that applies identically to regular and shifted windows and remains compatible with standard FlashAttention primitives. Our contribution is therefore not a new positional encoding itself, but using this formulation to make Swin fully trainable with fused attention and to study the resulting large-window, dense-token regime.

\textbf{Tokenization and spatial detail.} Tokenization determines how much fine-grained spatial structure enters the transformer backbone. Standard models like ViT and Swin embed non-overlapping patches ($p \times p$ convolution with stride $p$), creating hard borders where high-frequency edge information and local continuity are permanently lost~\cite{dosovitskiy2020image,liu2021swin}. Subsequent architectures adopt overlapping or convolutional tokenization to preserve local context across boundaries, as in PVTv2~\cite{wang2022pvt}. In our setting, this boundary fragmentation is exacerbated in the dense-token regime ($p=2$), where the number of patch borders quadruples. \name{} pairs window-local RoPE with an overlapping patch stem ($2p-1$ kernel, stride $p$, padding $p-1$), ensuring that both spatial position and high-frequency boundary detail survive the transition to fused attention and dense tokens.

%% file: sec/3_methods.tex
\section{Method}

While preserving Swin’s macro-hierarchy, \name{} updates three core components: it replaces standard windowed attention with FlashAttention to bypass score-matrix materialization, restores lost spatial awareness via a window-local learnable 2D RoPE on queries and keys(Figure~\ref{fig:token_arch_memory}B), and adopts overlapping token embeddings to preserve fine spatial structure (Figure~\ref{fig:token_arch_memory}A).

{\begin{figure*}[!thb]
    \centering
     \includegraphics[trim={0.0cm 11.5cm 0.0cm 0.0cm}, clip=true,width=0.90\linewidth]{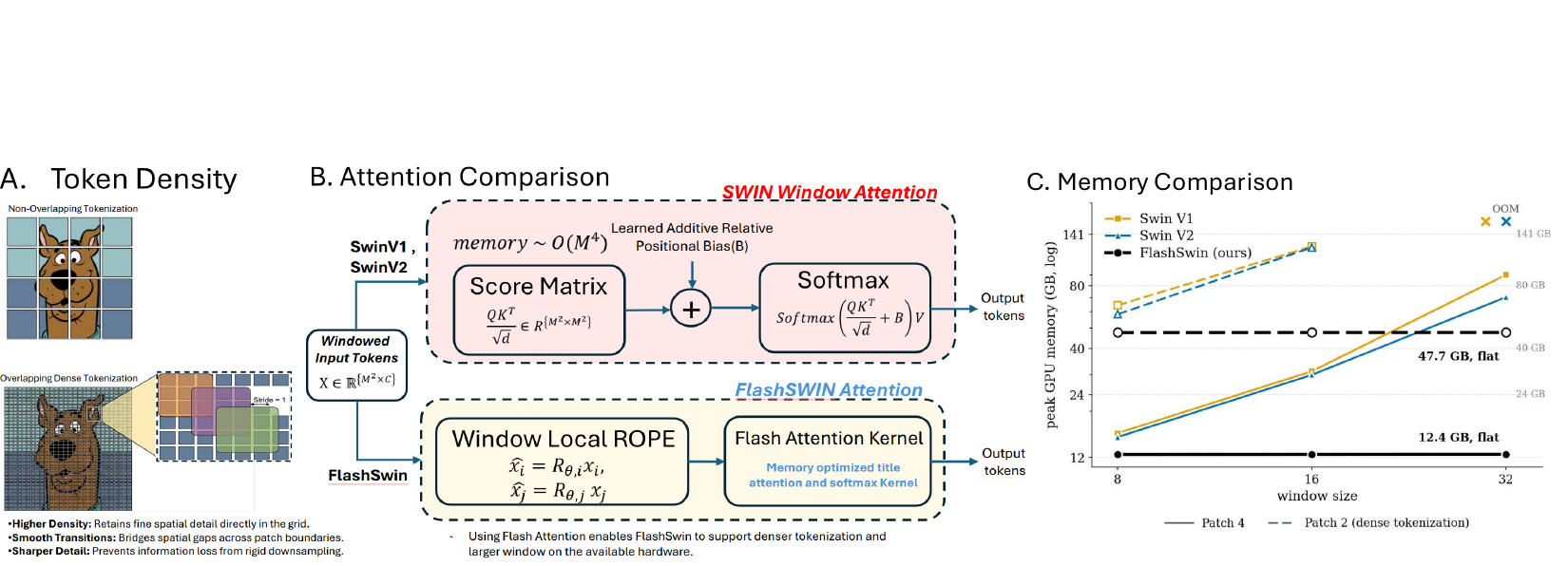}
  \caption{\textbf{Token density, architecture, and memory scaling comparison between Swin and \name.} (A) An overlapping stem ($P=2$) creates dense token grids that preserve spatial boundaries better than standard coarse patches ($P=4$). (B) \name{} replaces Swin's materialized score matrix and additive relative bias ($O(M^4)$ memory) with Window-Local 2D RoPE and a fused FlashAttention \cite{dao2022flashattention} kernel. (C) Peak training memory vs. window size $M$ on an H200 GPU ($256\times256$ resolution, batch size $128$). \name{} maintains a flat memory footprint as $M$ increases, whereas Swin-V1/V2 grow rapidly and run out of memory at dense settings.}
\label{fig:token_arch_memory}
\vspace{-1.5em}
\end{figure*}

\textbf{Memory-Efficient Windowed Attention.} Given an $H \times W$ token grid, hierarchical swin transformer partitions the feature map into non-overlapping $M \times M$ windows and applies self-attention independently within each window. With $N = M^2$ tokens per window,
\begin{equation}
\mathrm{Attn}(Q,K,V) = \mathrm{softmax}\!\left(
\frac{QK^\top}{\sqrt{d}} + B \right) V,
\label{eq:attn}
\end{equation}

where $B \in \mathbb{R}^{N \times N}$ is the learned relative-position bias. Standard attention materializes an $N \times N$ matrix per window, costing $O(N^2) = O(M^4)$ memory. Across $HW/M^2$ windows, total activation memory scales as $O(HW\,M^2)$, growing with both window size and token density, making their combination particularly expensive. \name{} instead evaluates attention within each window using an IO-aware fused kernel based on FlashAttention~\cite{dao2022flashattention}. The computation is tiled and the softmax is accumulated online, so the complete $N \times N$ matrix is never stored. This reduces auxiliary attention memory from $O(N^2)$ to $O(N)$ per window, equivalently from $O(M^4)$ to $O(M^2)$, and stage-level storage to approximately $O(HW)$, largely independent of window size. Only the implementation of attention changes: the same softmax attention is computed without approximation. While FlashAttention is widely used for long-sequence processing, its implications for scaling window size and token density in hierarchical vision transformers remain comparatively underexplored. The bias term $B$ in \cref{eq:attn} is the obstacle to this substitution, since it is added elementwise to the score matrix the fused kernel never forms. Removing it outright costs accuracy (\cref{tab:abl-comp}), so we instead encode position in the queries and keys.

\textbf{Window-Local Learnable RoPE.} We replace Swin's learned relative-position bias with a two-dimensional rotary position encoding (RoPE)~\cite{su2024roformer}, which injects positional information into the queries and keys rather than into the attention logits and is therefore compatible with the fused kernel. For each attention head, RoPE is applied over the full head dimension $d$, with channels split equally between the two spatial axes: the first half encodes the vertical coordinate $y$ and the second half the horizontal coordinate $x$. Both axes share a single learnable frequency set $\boldsymbol{\theta}$,
\begin{equation}
\widetilde{q}(x,y) =
\left[\, \mathcal{R}(y;\boldsymbol{\theta})\, q_{1:d/2} \;;\;
\mathcal{R}(x;\boldsymbol{\theta})\, q_{d/2+1:d} \,\right],
\label{eq:rope}
\end{equation}
with the same transformation applied to the keys. Here $\mathcal{R}(p;\boldsymbol{\theta})$ applies $2 \times 2$ rotations $R(p\theta_i)$ to consecutive channel pairs. The frequencies are parameterized as $\theta_i = \exp(\omega_i)$, where $\omega_i$ is initialized from the standard geometric RoPE schedule and optimized jointly with the network. Because the frequency set is shared across both axes, each head introduces only $d/2$ learned parameters.

Rotations compose as $\mathcal{R}(p)^\top \mathcal{R}(p') = \mathcal{R}(p' - p)$, so the inner product $\widetilde{q}(x,y)^\top \widetilde{k}(x',y')$ depends on the displacement $(x' - x,\, y' - y)$ alone. Relative spatial structure is thus preserved without an additive term, which is the property that makes \cref{eq:attn} expressible under a fused kernel.

We define these coordinates locally within each window. This matches the inductive bias of Swin, where attention is restricted to local windows and position is modeled through relative offsets within a window; global coordinates are less suitable because the operator never compares tokens across the full image. Under the shifted-window configuration, coordinates are recomputed relative to each window's own origin after the cyclic shift, so displacements remain window-local and the shift introduces no discontinuity in the encoding. Attention masking for wrapped windows is applied as an additive mask passed to the fused kernel, which supports masking without materializing the full score matrix.

Unlike Swin's discrete bias table, this parameterization is independent of the window size $M$. Swin stores $(2M-1)^2$ bias entries per head---$3969$ at $M{=}32$---defined only over the offsets seen during training, whereas \cref{eq:rope} uses $d/2$ frequencies that are continuous in the coordinate difference. The same learned frequencies can therefore be evaluated at any window size or input resolution without interpolation, which matters here because window size is varied explicitly as a scaling dimension.

\textbf{Overlapping Token Embedding.} Swin tokenizes the input with a non-overlapping patch projection, implemented as a convolution with kernel size and stride $p$. This creates hard boundaries between neighboring patches, separating local structures that cross a patch border before they enter the transformer. \name{} replaces this stem with an overlapping token embedding: a strided convolution with kernel size $2p-1$, stride $p$, and padding $p-1$. The stride is unchanged, so output resolution and token count are identical to the standard patch-$p$ embedding, but the larger kernel gives adjacent tokens overlapping receptive fields and allows them to share context across patch boundaries. The modification is most relevant in the dense-token regime, where a smaller patch preserves finer detail but also multiplies the boundaries at which it can be lost. The stem adds a lightweight local inductive bias while leaving the rest of the hierarchy unchanged.

\textbf{Resulting Design Space.} 
These three components are individually simple, yet jointly essential. Fused attention eliminates the $O(M^4)$ window-scaling bottleneck, but cannot be used alongside Swin's additive bias. Window-local RoPE provides positional awareness in a kernel-compatible, scale-invariant form. Finally, the overlapping stem mitigates spatial boundary fragmentation caused by finer tokenization. Removing any single component forfeits either memory savings or representation accuracy (\cref{tab:abl-comp}).While fused attention improves throughput, its key architectural impact is decoupling activation memory from window size $M$. While total token count remains bounded by spatial resolution, window expansion is no longer constrained by memory spikes. This allows \name{} to scale local receptive fields for substantially larger context modeling without memory overhead. The following experiments evaluate these expanded operating points ($M \in [8, 32]$, patch size $P \in [2, 4]$) across core vision tasks.

%% file: sec/4_results.tex
\section{Results and Discussion}
\label{sec:experiments}

\textbf{Experimental setup.} We evaluate \name{} on ImageNet-1K classification~\cite{russakovsky2015imagenet}, COCO detection~\cite{lin2014microsoft}, and ADE20K segmentation~\cite{zhou2019semantic}. Unless stated otherwise, all experiments use the Tiny configuration matched to Swin-T and SwinV2-T ($C{=}96$, depths $\{2,2,6,2\}$, heads $\{3,6,12,24\}$, ${\approx}28$M parameters); the four-stage hierarchy, patch merging, and alternating regular and shifted windows are unchanged from Swin. We vary window size $M \in \{8,16,32\}$ and patch size $p \in \{4,2\}$, training all models under the standard Swin recipe~\cite{liu2021swin} with matched schedules and optimization settings. The SwinV1/V2-T baselines are re-trained (locally) at $256^2$ under this identical recipe. All models, including baselines, are trained under PyTorch AMP autocast (\textit{fp16}). \name{}'s fused attention consumes \textit{bf16} queries, keys, and values, applies the rotary embedding in \textit{fp32}, and accumulates the softmax in \textit{fp32} inside the kernel, returning an \textit{fp32} output without ever materializing the $N\!\times\!N$ attention matrix. The Swin V1/V2 baselines compute $A=QK^{\top}$ and $AV$ in \textit{fp16} with the softmax promoted to \textit{fp32} by autocast, and must materialize the attention matrix because their relative position bias is an additive term on it. Peak memory for the models is measured over a full forward and backward pass at batch size $128$ and $256{\times}256$ input on an H200, without gradient checkpointing, unless noted otherwise. Classification models are trained on four H200 GPUs, and downstream models for segmentation and detection are trained on four and two A100-40GB GPUs, respectively. We additionally compare against the trainable low-rank formulation of FlashBias~\cite{wu2026flashbias} to evaluate an alternative approach that retains additive positional bias within fused attention. Additional implementation details are provided in \Cref{ssec:additional_implementation_details}.

\textbf{Robustness.} We evaluate the trained IN-1K models without fine-tuning on ImageNet-A~\cite{hendrycks2021natural}, -R~\cite{hendrycks2021many}, and -Sketch~\cite{wang2019learning} (top-1 over the standard $200$-class subsets for A and R), and ImageNet-C~\cite{hendrycks2019benchmarking} (AlexNet-normalized mean corruption error, mCE, over $15$ corruptions at $5$ severities). We evaluate the released $224^2$ ImageNet-C images with the standard transform rather than regenerating corruptions at $256^2$; the resulting $1.30\times$ upsampling makes absolute mCE incomparable to the $224$ protocol but is applied identically to every model.

\textbf{Downstream transfer.} For semantic segmentation, we evaluate UPerNet~\cite{xiao2018unified} on ADE20K. Beyond standard mIoU, we report Boundary-IoU~\cite{cheng2021boundary} and boundary F1~\cite{csurka2013good} near object contours, where larger windows and denser tokens exert the greatest impact. For object detection, we evaluate Mask R-CNN~\cite{he2017mask} on COCO~\cite{lin2014microsoft}. To keep the dense $p{=}2$ setting tractable, inputs are uniformly padded to $512{\times}512$ rather than $1333{\times}800$. Window size $M$ is fixed to its pretraining value throughout, meaning larger inputs increase window count rather than window scale. Full implementation and training details are provided in the Supplementary \ref{ssec:additional_implementation_details}.

\subsection{Memory Scaling}
\label{sec:memory}

\Cref{fig:token_arch_memory}C, \Cref{fig:throughput}, and \Cref{tab:memory} show the central efficiency result: \name{} activation memory is largely decoupled from window size, and so is throughput. At patch size $4$, \name-T remains at $12.4$\,GB as $M$ grows from $8$ to $32$, while Swin-T increases from $15.7$ to $90.1$\,GB ($5.7\times$) and SwinV2-T from $15.0$ to $70.4$\,GB. The baselines pay for larger windows in both memory and speed: Swin-T falls from $1173$ to $442$ img/s and SwinV2-T from $871$ to $371$, whereas \name-T remains near $1$K img/s. At $M{=}8$, Swin-T is slightly faster because the small attention matrix does not amortize fused-kernel overhead; \name{} overtakes both baselines from $M{=}16$ and is $2.4\times$ faster than Swin-T at $M{=}32$. FlashBias is likewise flat in $M$, confirming that fusion, rather than the positional encoding, removes the window dependence; however, \name{} uses less memory, becomes faster at larger windows, and remains fixed at $28.3$M parameters while FlashBias grows with $M$.

Denser grids move the baselines closer to the hardware ceiling. At patch size $2$, \name-T holds at $47.7$\,GB and ${\sim}290$ img/s across windows, while both Swin variants exceed $120$\,GB at $M{=}16$. At $M{=}32$, Swin-T cannot train even with checkpointing~\cite{chen2016training}, and SwinV2-T runs only with checkpointing at $138.9$\,GB and $57$ img/s, against \name{}'s $47.7$\,GB and $272$ img/s without it---$2.9\times$ less memory at $4.8\times$ the throughput. FlashBias\cite{wu2026flashbias} remains memory-stable at $51.5$\,GB but reaches only $205$ img/s, while its $M{=}8$ configuration cannot launch at batch $128$ because of its kernel grid limit. Checkpointing further reduces memory but trades it for recomputation, making the ability to train directly consequential for throughput. Finally, the $4\times$ increase in token count from patch $4$ to $2$ costs \name{} only $3.85\times$ the memory ($12.4\!\to\!47.7$\,GB), while increasing $M$ adds essentially none. Thus, \name{} shifts the memory scaling from window size to token count: standard windowed attention becomes memory-bound long before large windows cease to be useful, and that is the constraint \name{} removes.

\subsection{ImageNet-1K Classification Results}

\Cref{tab:cls} reports \name-T across progressively larger windows and denser token grids. At matched compute ($M{=}8$, $p{=}4$), \name-T reaches $82.3\%$ top-1, outperforming both Swin baselines at the same $5.9$ GFLOPs. Increasing the window to $M{=}16$ raises accuracy to $82.9\%$ and improves ImageNet-V2 from $70.8\%$ to $71.9\%$. Classification then saturates: increasing $M$ to $32$ adds $27\%$ FLOPs without improving ImageNet-1K and yields only $+0.1$ on ImageNet-V2. Larger windows therefore help up to a point, with their benefit more visible on the held-out ImageNet-V2 set than on clean ImageNet---a pattern that becomes stronger under distribution shift.

The largest gain comes from token density. At $M{=}32$ and $p{=}2$, \name-T reaches $84.1\%$ ImageNet-1K and $73.4\%$ ImageNet-V2, improving by $1.8$ and $2.6$ points over the $M{=}8$, $p{=}4$ \name-T configuration while retaining the same ${\approx}28$M model capacity. Thus, accuracy is governed not only by parameter count, but also by the spatial resolution at which those parameters operate: $84.1\%$ at ${\approx}28$M approaches Base-scale accuracy (${\sim}84.5\%$~\cite{liu2022swin}) with substantially fewer parameters. Higher token density does increase compute ($37.4$ vs.\ $5.9$ GFLOPs), but standard windowed attention cannot train this configuration at the same batch size without activation checkpointing~\cite{chen2016training}, whose recomputation cost \name{} avoids.

Fusion alone does not explain the accuracy gains. FlashBias-T\cite{wu2026flashbias} retains an additive relative-position bias through a learned rank-$16$ factorization compatible with fused attention. At $M{=}32$, $p{=}4$, it reaches $82.7\%$, within $0.2$ points of \name-T, but uses $29.5$M parameters versus $28.3$M because its factors grow with $M^2$. On clean ImageNet, the two are therefore nearly indistinguishable. Their difference becomes substantially clearer under distribution shift and downstream transfer.

\begin{table}[t]
\centering
\small
\setlength{\tabcolsep}{2pt}
\begin{tabular}{lccccccc}
\toprule
Method & Params & $M$ & $p$ & Res & GFLOPs & IN-1K$\uparrow$  & IN-V2$\uparrow$ \\
\midrule
\multicolumn{8}{l}{\textit{Matched compute}} \\
Swin-T      & 28M & 8  & 4 & 256 & 5.9  & 81.9           & 70.2           \\
SwinV2-T    & 28M & 8  & 4 & 256 & 5.9  & 81.9           & 70.3           \\
\name-T     & 28M & 8  & 4 & 256 & 5.9  & \textbf{82.3} & \textbf{70.8} \\
\midrule
\multicolumn{8}{l}{\textit{Large-window}} \\
SwinV2-T    & 28M & 16 & 4 & 256 & 6.6  & 82.8           & 71.4           \\
\name-T     & 28M & 16 & 4 & 256 & 6.6  & \textbf{82.9} & 71.9 \\
FlashBias-T & 30M & 32 & 4 & 256 & 9.1  & 82.7           & 71.3             \\
\name-T     & 28M & 32 & 4 & 256 & 8.4  & \textbf{82.9}  & \textbf{72.0}           \\
\midrule
\midrule
\multicolumn{8}{l}{\textit{Dense-token regime}}  \\
\name-T     & 28M & 32 & 2 & 256 & 37.4 & \textbf{84.1} & \textbf{73.4} \\
%\name-T     & 28M & 32 & 4 & 512 & 37.4 & 84.0           & 72.9           \\
\bottomrule
\end{tabular}
  \vspace{-0.5em}
\caption{\textbf{ImageNet-1K classification.} $M$ is window size, $p$ patch size; accuracy is top-1 (\%) on ImageNet-1K and on the held-out ImageNet-V2 set. Only the first block is compute-matched; later blocks trade GFLOPs for window size and token density at matched parameter count, the exception being FlashBias-T, whose low-rank factors scale with $M^2$. \textbf{Bold}: best in block.}
\label{tab:cls}
 \vspace{-1.5em}
\end{table}

\textbf{Out-of-distribution robustness.}
We evaluate on four distribution-shift benchmarks (\cref{tab:robustness}). At $M{=}8$, \name-T already outperforms both Swin variants on every benchmark, with larger margins than on clean ImageNet. Increasing the window to $M{=}16$ further improves robustness at fixed token density, while dense tokenization produces the largest gains: at $M{=}32$, $p{=}2$, \name-T reaches $42.8\%$ ImageNet-A, $46.6\%$ ImageNet-R, $33.0\%$ ImageNet-Sketch, and $50.7$ mCE. Relative to SwinV2-T, the ImageNet-A improvement grows from $+3.4$ points at matched configuration to $+18.1$ points in the dense-token regime, compared with only $+2.2$ points on clean ImageNet. Thus, the larger-window, dense-token regime has a disproportionately large benefit under distribution shift.

FlashBias provides an important control. Despite nearly matching \name-T on clean ImageNet at $M{=}32$ ($82.7\%$ vs.\ $82.9\%$), it falls behind on natural shifts: by $4.5$ points on ImageNet-A, $2.6$ on ImageNet-R, and $3.0$ on Sketch. The gap is smaller on synthetic corruption, where FlashBias reaches $54.7$ mCE versus \name-T's $53.0$.

\begin{table}[t]
\setlength{\tabcolsep}{1pt}
  \centering
  \small
  \begin{tabular}{lcccccc}
  \toprule
  Backbone & M/p/Res & IN-1K$\uparrow$  & IN-A$\uparrow$  & IN-R$\uparrow$  & Sketch$\uparrow$  & IN-C$\downarrow$ \\
  \midrule
  Swin-T      & 8/4/256   & 81.9 & 25.1 & 41.7 & 28.0 & 58.5 \\
  SwinV2-T    & 8/4/256   & 81.9 & 24.7 & 41.3 & 27.9 & 58.3 \\
  \name-T     & 8/4/256   & 82.3 & \textbf{28.1} & \textbf{43.6} & \textbf{30.4} & \textbf{56.9} \\
  \midrule
  SwinV2-T    & 16/4/256  & 82.8 & 31.2 & 44.6 & 31.4 & 54.1 \\
  \name-T     & 16/4/256  & 82.9 & \textbf{33.2} & \textbf{45.6} & \textbf{32.9} & 53.7 \\
  FlashBias-T & 32/4/256  & 82.7 & 28.4 & 43.0 & 29.4 & 54.7 \\
  \name-T     & 32/4/256  & 82.9 & 32.9 & \textbf{45.6} & 32.4 & \textbf{53.0} \\
  \midrule
  \midrule
  \name-T     & 32/2/256  & 84.1 & \textbf{42.8} & \textbf{46.6} & \textbf{33.0} & \textbf{50.7} \\
  \bottomrule
  \end{tabular}
  \vspace{-0.5em}
\caption{\textbf{Out-of-distribution robustness.} Top-1 (\%) on ImageNet-A, -R, and -Sketch; mean corruption error (mCE) on ImageNet-C. IN-1K repeats clean accuracy for reference and is excluded from bolding; \textbf{bold}: best in each window block. ImageNet-A and Sketch peak at $M{=}16$ while mCE keeps falling; dense tokenization is strongest on all four. FlashBias-T fuses the kernel but keeps a rank-$16$ additive bias: matched to \name-T on clean accuracy, it scores at the $M{=}8$ level on natural shift despite $M{=}32$, while remaining competitive on synthetic corruption.}
  \label{tab:robustness}
  \vspace{-1.5em}
  \end{table}

\textbf{Component ablation.} \Cref{tab:abl-comp} and \cref{fig:ablation-curves} isolate the components needed to integrate memory-efficient attention into Swin. Naively using a fused kernel—dropping Swin's relative-position bias—drops ImageNet-1K accuracy to $80.0\%$ ($1.9$ points below SwinV2-T). Because fused attention is exact, this drop stems entirely from lost intra-window positional signal. Learnable window-local RoPE recovers $1.8$ points ($81.8\%$), matching SwinV2-T ($81.9\%$) and proving kernel compatibility costs no accuracy. The overlapping stem adds $+0.5\%$, bringing full \name-T to $82.3\%$. Training curves (\cref{fig:ablation-curves}) confirm this ranking is established early and maintained throughout. RoPE and the stem provide a $+2.3$ point boost over naive fused attention, turning memory savings into performance gains.

\subsection{ADE20K Semantic Segmentation}
All backbones follow the UPerNet protocol described above. In \cref{tab:seg}, the third value in M/p/Res denotes the backbone pretraining resolution. Boundary-IoU~\cite{cheng2021boundary} uses a $2\%$ diagonal band and boundary F1 (BF)~\cite{csurka2013good} a $0.8\%$ matching tolerance; both are accumulated per class over the full $2{,}000$-image validation set and reduced across all $150$ classes, following mIoU.

At matched windows, \name-T outperforms all baselines across all three metrics. At $M{=}8$, it improves over SwinV2-T by $0.92$ mIoU, $0.85$ Boundary-IoU, and $1.19$ BF. Increasing $M$ to $16$ and $32$ further improves all three, reaching $46.88/35.55/49.72$ at $M{=}32$. Unlike classification, which saturates beyond $M{=}16$, segmentation continues to benefit from larger windows, indicating that broader spatial context remains useful for dense prediction. At the same $M{=}32$, FlashBias-T trails \name-T by $1.73$ mIoU, $1.78$ Boundary-IoU, and $1.42$ BF despite nearly matching it on ImageNet, again showing that fusion alone does not reproduce the downstream gains of \name{}.

\textbf{Dense tokenization improves boundaries; wider windows improve both.}
At fixed $M{=}32$, reducing patch size from $4$ to $2$ improves mIoU by $0.40$ points, but Boundary-IoU by $0.87$ and BF by $1.34$---roughly $2$--$3\times$ the mIoU gain. In contrast, widening the window from $8$ to $32$ at fixed density adds $2.49$ mIoU, $2.44$ Boundary-IoU, and $2.36$ BF. Thus, broader context improves region and boundary quality together, whereas denser tokens disproportionately improve boundary localization. The BF tolerance sweep (\cref{tab:bf-tolerance}) supports this distinction: the dense configuration gains most under strict tolerances, with its margin decreasing as the tolerance is relaxed. At the same ${\approx}28$M capacity, the dense $M{=}32$ model gains $+4.54$ Boundary-IoU and $+5.43$ BF over Swin-T, demonstrating the value of the high-density regime enabled by memory-efficient attention.

\begin{table}[t]
  \centering
  \small
  \setlength{\tabcolsep}{4pt}
  \begin{tabular}{lcccc}
  \toprule
  Backbone & M/p/Res & mIoU$\uparrow$ & Boundary-IoU$\uparrow$ & BF$\uparrow$ \\
  \midrule
  Swin-T      & 8/4/256   & 43.20 & 31.88 & 45.63 \\
  SwinV2-T    & 8/4/256   & 43.47 & 32.26 & 46.17 \\
  \name-T     & 8/4/256   & \bf 44.39 & \bf 33.11 & \bf 47.36 \\
  \midrule
  SwinV2-T    & 16/4/256  & 45.46 & 34.14 & 48.22 \\
  \name-T     & 16/4/256  & 46.03 &  34.73 & 48.98 \\
  FlashBias-T & 32/4/256  & 45.15 & 33.77 & 48.30 \\
  \name-T     & 32/4/256  & \bf 46.88 & \bf 35.55 & \bf 49.72 \\
  \midrule
  \midrule
  \name-T     & 32/2/256  & \bf 47.28 & \bf 36.42 & \bf 51.06 \\
  \bottomrule
  \end{tabular}
  \vspace{-0.5em}
  \caption{\textbf{ADE20K segmentation and boundary quality.} Single-scale mIoU, Boundary-IoU, and F1. \name{}-T leads across all matched windows. Larger windows improve region and boundary metrics equally, while token density ($p{=}2$) yields $2\text{--}3\times$ larger gains on boundaries than mIoU. FlashBias-T, matched to \name-T on ImageNet, trails it by $1.7$ mIoU and by a similar relative margin on both boundary metrics, performing at SwinV2-T's $M{=}16$ level despite $M{=}32$. \textbf{Bold}: best in block.}
  \label{tab:seg}
\vspace{-1.5em}
\end{table}

\subsection{COCO Object Detection}
All backbones are evaluated on COCO using Mask R-CNN under the matched $512\times512$ protocol above. At $M{=}8$, \name-T improves over both baselines, reaching $38.5$ box AP and $35.3$ mask AP versus $37.3/34.4$ for SwinV2-T. Increasing $M$ to $16$ raises performance to $39.0/36.0$, and $M{=}32$ further improves it to $40.3/36.7$. Unlike classification, which saturates beyond $M{=}16$ (\cref{tab:cls}), detection continues to benefit from larger windows, suggesting that broader spatial context is particularly useful for localization and instance delineation. In contrast, FlashBias-T at $M{=}32$ reaches only $37.0/34.3$ AP despite nearly matching \name-T on ImageNet, placing it near the $M{=}8$ Swin baseline. Thus, the downstream benefit of larger windows is not explained by fused attention alone.

Reducing the patch size to $p{=}2$ at $M{=}32$ produces the largest gain: $44.1$ box AP and $39.9$ mask AP, improving over the $p{=}4$ model by $3.8$ and $3.2$ points and over Swin-T by $7.0$ and $5.7$. The gain is concentrated on small objects: AP$_S$ increases from $20.1$ to $27.1$ for boxes and from $14.4$ to $19.1$ for masks, while box AP$_L$ remains unchanged at $59.7$ (\cref{ssec:detection_by_size}). This links dense tokenization to where spatial resolution is most limiting: small objects represented by only a few tokens under patch-$4$. Together with the ADE20K boundary results, this shows that denser tokenization is particularly beneficial when prediction depends on fine spatial detail.

\begin{table}[!htb]
\centering
\small
\setlength{\tabcolsep}{2pt}
\begin{tabular}{lccccccc}
\toprule
& & \multicolumn{3}{c}{Box$\uparrow$} & \multicolumn{3}{c}{Mask$\uparrow$} \\
\cmidrule(lr){3-5}\cmidrule(lr){6-8}
Backbone & M/p/Res & AP & AP$_{50}$ & AP$_{75}$ & AP & AP$_{50}$ & AP$_{75}$ \\
\midrule
Swin-T      & 8/4/256  & 37.1 & 58.2 & 40.1 & 34.2 & 54.9 & 36.2 \\
SwinV2-T    & 8/4/256  & 37.3 & 58.8 & 40.3 & 34.4 & 55.3 & 36.4 \\
\name-T     & 8/4/256  & \bf 38.5 & \bf 59.8 & \bf 41.4 & \bf 35.3 & \bf 56.8 & \bf 37.6 \\
\midrule
SwinV2-T    & 16/4/256 & 39.0 & 60.3 & 42.2 & 35.7 & 57.0 & 37.9 \\
\name-T     & 16/4/256 & 39.0 & 60.2 & 42.2 & \bf 36.0 & \bf 57.3 & \bf 38.1 \\
FlashBias-T & 32/4/256 & 37.0 & 58.2 & 39.7 & 34.3 & 55.3 & 36.6 \\
\name-T     & 32/4/256 & \bf 40.3 & \bf 61.7 & \bf 43.6 & \bf 36.7 & \bf 58.6 & \bf 39.3 \\
\midrule
\midrule
\name-T     & 32/2/256 & \bf 44.1 & \bf 66.6 & \bf 48.2 & \bf 39.9 & \bf 63.3 & \bf 42.4 \\
\bottomrule
\end{tabular}
\vspace{-.5em}
\caption{\textbf{COCO detection and instance segmentation.} \name-T improves with window
size and matches or beats the Swin baselines at each matched window; halving the patch at
$M{=}32$ gives the largest gain. \textbf{FlashBias-T} fuses the kernel but keeps an additive
bias as a learned rank-16 factorization: it ties \name{} on ImageNet ($82.65$ vs $82.94$)
yet loses $3.3$ box AP, performing at Swin-T's $M{=}8$ level despite $M{=}32$.}
\label{tab:detection}
\vspace{-1.5em}
\end{table}

\subsection{Large Model Experiments}

To test whether the benefits of \name{} persist with increasing model capacity, we scale \name-L to approximately $197$M parameters, closely matching Swin-L \cite{liu2021swin} and SwinV2-L \cite{liu2022swin}. For the baselines, we use the official IN-22K pretrained checkpoints and perform downstream training under the same evaluation pipeline. \name-L is pretrained on IN-22K \cite{ridnik2021imagenet} at $192^2$ with patch size $2$ and window size $32$ for 90 epochs. At the same $192^2$ input resolution, this produces a $96^2$ stage-1 token grid compared with $48^2$ for SwinV2-L, corresponding to $4\times$ more tokens. The models are therefore parameter-matched but not compute-matched; this comparison specifically tests whether the denser-token regime enabled by \name{} remains beneficial as model capacity increases.

\textbf{Classification.}  On IN-1K the three Large models are closely matched: \name-L reaches $86.91\%$, within $0.02$ of SwinV2-L ($86.93\%$) and $0.40$ above Swin-L ($86.51\%$). The ordering changes on ImageNet-V2, a fresh sample of the same distribution: \name-L leads at $78.02\%$ against $77.48\%$ and $76.81\%$, and degrades least from clean accuracy ($-8.89$ points against $-9.45$ and $-9.70$). With essentially identical IN-1K accuracy, the higher ImageNet-V2 result indicates better generalization to the independently collected test set. \name-L is likewise the strongest of the three on ImageNet-A and -R; the full out-of-distribution comparison is reported in the supplementary \ref{ssec:out_distribution_large}.

\textbf{Segmentation.} On ADE20K, \name-L achieves $52.58$ mIoU, improving over Swin-L by $+2.45$ points and SwinV2-L by $+4.23$ points. The improvement is even more pronounced for boundary quality: relative to Swin-L, \name-L gains $+3.02$ Boundary-IoU and $+2.98$ BF. In relative terms, Boundary-IoU improves by $7.8\%$ compared with $4.9\%$ for mIoU, reproducing the pattern observed at the Tiny scale (\cref{tab:seg}): denser spatial representations disproportionately benefit boundary localization. Importantly, these gains are obtained without increasing parameter count. Instead, \name-L uses its improved memory scaling to process a substantially denser token grid and larger attention windows. Together with the Tiny-scale results, this suggests that scaling vision transformers is not solely a question of increasing model parameters; increasing spatial resolution available to a fixed-capacity backbone provides an additional and complementary scaling axis. A per-class decomposition (Refer \ref{sec:bf-tolerance}) shows the gain is broad rather than category-specific: $41$ of $45$ well-supported classes improve, with the largest gains on finely contoured objects and the only losses on diffuse regions.
\begin{table}[!htb]
\centering
\small
\setlength{\tabcolsep}{1pt}
\begin{tabular}{lcccccccc}
\toprule
Backbone & $p$ & $M$ & PT Res. & IN-1K & IN-V2 & mIoU & B-IoU & BF \\
\midrule
Swin-L   & 4 & 7  & $224^2$ & 86.51 & 76.81 & 50.13 & 38.73 & 54.31 \\
SwinV2-L & 4 & 12 & $192^2$ & \bf 86.93 & 77.48 & 48.35 & 37.54 & 52.72 \\
\name-L  & 2 & 32 & $192^2$ & 86.91 & \bf 78.02 & \bf 52.58 & \bf 41.75 & \bf 57.29 \\
\bottomrule
\end{tabular}
\caption{\textbf{Large-scale results.} PT Res.\ denotes IN-22K pretraining resolution; all IN-1K/IN-V2 results use $256^2$ fine-tuning. \name-L uses $4\times$ more stage-1 tokens than SwinV2-L at the same $192^2$ pretraining resolution. Published Swin-L/SwinV2-L results at $384^2$ fine-tuning are not directly comparable. Higher is better for all metrics.}
\label{tab:large}
\end{table}

\subsection{Token Density via finer patches vs. higher res.} 
Because \name decouples memory from window size, a target token density can be reached two ways: a finer patch embedding at low input resolution, or a coarser patch at high resolution. Table~\ref{tab:density-route} contrasts the two at a fixed $128\times128$ token grid with matched FLOPs and parameters. The two routes are nearly interchangeable on these benchmarks. They are near-identical in what they learn ($84.1$ vs.\ $84.0$ top-1, with the finer patch $+0.5$ ahead on ImageNet-V2) and, at equal batch size, near-identical in what they cost: $24.4$ vs.\ $23.8$GB of training memory and $287$ vs.\ $284$ images/s. This is expected-beyond the first reduction the two networks are architecturally identical, and the stem is too small a fraction of activations to separate them-but it is worth establishing, because it isolates token count as the variable that matters. 

We state the conclusion narrowly. ImageNet-1K has a median short side of $374$ pixels and only $6.4\%$ of images reach $512$, so the coarse-patch arm is trained largely on upsampled data. The experiment therefore shows that, at these resolutions, a coarse stem on an interpolated image can match a finer stem on the original; it does not establish that native high-resolution input is unnecessary. The more consequential setting is data whose task-relevant signal exists at the native pixel scale, such as cellular morphology in histopathology or small lesions and thin structures in medical imaging~\cite{li2026pants,wu2024voco,baroni2024optimizing,dede2025deep}. Evaluating \name{} in these genuinely high-resolution regimes is an important direction for future work. 

%% file: sec/5_conclusion.tex
\section{Conclusion and Future Work}
In this work, we present \name{}, an architecture that removes the window-size memory wall in hierarchical vision transformers. By replacing score-matrix relative-position bias with window-local learnable 2D RoPE, \name{} enables standard FlashAttention within windowed self-attention and makes previously impractical large-window, dense-token regimes ($M{=}32$, $p{=}2$) trainable. Our comparison with FlashBias\cite{wu2026flashbias} further shows that fused attention itself is sufficient to flatten memory scaling with window size, but not to reproduce the downstream gains of \name{}. Across classification, detection, segmentation, and distribution shift, larger windows improve contextual modeling while denser tokens provide particularly strong gains in fine spatial localization. These benefits persist from Tiny ($28$M) to Large ($197$M) models, with \name-L reaching $52.58$ mIoU on ADE20K and outperforming both Swin-L and SwinV2-L. Together, our results show that scaling vision transformers is constrained not only by parameter count, but also by the spatial resolution that hardware memory allows them to process.

\textit{Limitations and future work.}
Although \name{} removes the dominant memory dependence on window size, dense tokenization still increases computation; the $p{=}2$ configuration therefore trades additional FLOPs for spatial resolution. Our current benchmarks also operate at relatively modest image resolutions and do not fully test the regime that motivates native high-resolution processing. Extending \name{} to gigapixel histopathology, high-resolution medical imaging~\cite{li2026pants,wu2024voco,baroni2024optimizing,dede2025deep}, and other domains where downsampling removes task-relevant detail is therefore an important direction. Dense-token, large-window backbones may also benefit vision-language models requiring fine-grained visual grounding and text recognition. Finally, extending memory-efficient windowed attention to 3D medical volumes and video is particularly promising, where the growth of volumetric attention makes the memory constraint even more severe.

\section*{Acknowledgments}
This work was supported by the National Science Foundation under Grant NSF2217154. We also gratefully acknowledge the Center for High Performance Computing at the University of Utah for providing the computational resources used in this work. These resources were partially supported by NIH Shared Instrumentation Grant 1S10OD021644-01A1. The content is solely the responsibility of the authors and does not necessarily represent the official views of the National Science Foundation or the National Institutes of Health.

%% file: sec/6_supp.tex
\setcounter{section}{0}
\setcounter{figure}{0}
\setcounter{table}{0}
\section{Supplementary}
\renewcommand{\thesection}{S\arabic{section}}
\renewcommand{\thefigure}{S\arabic{figure}}
\renewcommand{\thetable}{S\arabic{table}}

\begin{table*}[!htb]
\centering
\small
\setlength{\tabcolsep}{4pt}
\begin{tabular}{ll ccc c ccc c}
\toprule
& & \multicolumn{3}{c}{Memory (GB) $\downarrow$} & & \multicolumn{3}{c}{Train throughput (img/s) $\uparrow$} & \\
\cmidrule(lr){3-5}\cmidrule(lr){7-9}
$p$ & Model & $M{=}8$ & $M{=}16$ & $M{=}32$ & & $M{=}8$ & $M{=}16$ & $M{=}32$ & Params (M) \\
\midrule
\multirow{5}{*}{4}
 & Swin-T          & 15.7/6.4  & 31.0/11.6 & 90.1/37.7 & & \bf 1173/893 & 828/612 & 442/314 & 28.5 \\
 & SwinV2-T        & 15.0/5.4  & 29.8/12.0 & 70.4/34.9 & & 871/670 & 643/484 & 371/272 & 28.4 \\
 %& RoPE-Swin-T     & 17.8/6.8  & 32.3/11.8 & 87.4/38.4 & & 1096/836 & 863/647 & 528/387 & 28.3 \\
 & FlashBias-T     & 13.2/5.2  & 13.2/5.1  & 13.2/5.2  & & 934/741 & 924/733 & 826/657 & 28.5--29.5 \\
 & \bf \name-T     & \bf 12.4/4.2 & \bf 12.4/4.8 & \bf 12.4/5.0 & & 1058/815 & \bf 1103/850 & \bf 1043/808 & \bf 28.3 \\
\midrule
\multirow{5}{*}{2}
 & Swin-T          & 64.4/23.2 & 123.3/45.6 & OOM/OOM & & \bf 317/239 & 210/153 & OOM/OOM & 28.5 \\
 & SwinV2-T        & 58.3/19.8 & 122.0/48.7 & OOM/138.9 & & 235/180 & 165/123 & OOM/57 & 28.4 \\
 %& RoPE-Swin-T     & 71.3/25.6 & 130.9/46.1 & OOM/OOM & & 295/224 & 217/160 & OOM/OOM & 28.3 \\
 & FlashBias-T     & $\dagger$ & 51.4/19.1 & 51.5/19.2 & & $\dagger$ & 243/192 & 205/162 & 29.4--31.6 \\
 & \bf \name-T     & \bf 47.7/15.9 & \bf 47.7/15.9 & \bf 47.7/16.0 & & 289/221 & \bf 297/227 & \bf 272/210 & \bf 28.3 \\
\bottomrule
\end{tabular}
\caption{\textbf{Peak training memory and throughput} without/with gradient checkpointing, measured at batch size $128$ with $256^2$ inputs on one H200. FlashBias-T uses a trainable rank-$16$ bias factorization. Both FlashBias-T and \name-T remove the memory dependence on window size, while \name-T uses less memory and higher throughput at $M\!\geq\!16$. FlashBias parameters increase with $M$, whereas \name-T remains fixed at $28.3$M. $\dagger$ denotes a configuration that FlashBias cannot launch at batch $128$ due to its CUDA grid-size limit.}
\label{tab:memory}
\end{table*}

\begin{table}[!thb]
 \vspace{-0.5em}
\centering
\small
\begin{tabular}{lccc}
\toprule
Variant & RoPE & Overlap & IN-1K$\uparrow$ \\
\midrule
Naive Flash            & --         & --         & 80.0 \\
\quad + learnable RoPE & \checkmark & --         & 81.8 \\
\quad + overlap stem   & \checkmark & \checkmark & \textbf{82.3} \\
\midrule
\textit{SwinV2-T (ref.)} & \multicolumn{2}{c}{\textit{relative-position bias}} & \textit{81.9} \\
\bottomrule
\end{tabular}
\caption{\textbf{Component ablation} at $M{=}8$, $p{=}4$, $256^2$. Dropping Swin's relative-position bias by applying the fused FlashAttention kernel costs $1.9$ points against SwinV2-T.}
\label{tab:abl-comp}
 \vspace{-1.5em}
\end{table}

\subsection{Additional Implementation Details}\label{ssec:additional_implementation_details}

\paragraph{Effective window per stage for \name}. The window size $M$ reported throughout is the stage-1 value. When a stage's spatial resolution is not larger than $M$, the implementation clamps the window to that resolution and disables the shift, since a window larger than the feature map has no meaning. At $256^2$ input with $p{=}4$ the stage grids are $64$, $32$, $16$, $8$, so $M{=}32$ is realized as $32, 32, 16, 8$ across the four stages, and $M{=}64$ at $512^2$ is realized as $128 \rightarrow 64$ only from stage~2 onward. The deepest stages therefore perform global attention within their own resolution. This affects all three architectures identically and matches Swin's own convention, but it must be reproduced for the reported numbers to match: an implementation that instead pads to a multiple of $M$ will give different results at large $M$.

\paragraph{ADE20K Segmentation.} Experiments are implemented in \textit{mmsegmentation}~\cite{mmseg2020} using UPerNet~\cite{xiao2018unified} initialized from ImageNet-1K checkpoints. We train using the standard $160$K schedule with $512{\times}512$ crop sizes. Optimization uses AdamW with a learning rate of $1.2{\times}10^{-4}$ and weight decay of $0.01$. The effective batch size is $32$ ($4$ GPUs $\times$ $4$ images/GPU with $2$ gradient-accumulation steps).

\paragraph{COCO Detection.} Experiments are implemented in \textit{mmdetection}~\cite{mmdetection} using Mask R-CNN~\cite{he2017mask} under the standard $1\times$ schedule. Optimization uses AdamW with a learning rate of $10^{-4}$, weight decay of $0.05$, a $1000$-iteration warmup, and step decay at epochs $8$ and $11$. The effective batch size is $16$ ($2$ GPUs $\times$ $4$ images/GPU with $2$ gradient-accumulation steps). Inputs are padded to $512{\times}512$ across all models for uniform evaluation; absolute AP scores reflect this reduced input resolution.

\subsection{Boundary tolerance sensitivity}
\label{sec:bf-tolerance}
The BF-score depends on a matching tolerance $\tau$, expressed as a fraction of the image diagonal $d$. Any single choice of $\tau$ is arbitrary, and an improvement visible at one tolerance is not by itself evidence of better boundary quality. We therefore sweep $\tau/d \in \{0.002, 0.004, 0.008, 0.016, 0.032\}$ over the full 2{,}000-image ADE20K validation set, holding the models, the predictions, and the Boundary-IoU band fixed. Table~\ref{tab:bf-tolerance} reports the result.

The ordering of the seven backbones is identical at all five tolerances, so the boundary result is a property of the models rather than of the operating point. Beyond this invariance, the margin has a systematic shape: \name-T at $32/2$ leads the next-best model by $+1.50$ BF at $\tau/d{=}0.004$ and by only $+0.84$ at the loosest tolerance, decaying as $\tau$ grows past the strictest settings. A larger $\tau$ relaxes the criterion toward ordinary region overlap, so a margin that shrinks with $\tau$ is what one expects when the improvement comes from more precise boundary localization rather than from better region classification. Consistently, the patch-size reduction that yields only $+0.40$ mIoU yields $+0.87$ Boundary-IoU and $+1.34$ BF at $\tau/d = 0.008$ --- two to three times as much gain at the boundary as in aggregate region accuracy, detail that mIoU alone substantially under-reports.

\paragraph{Implementation.}
Both metrics accumulate per-class numerators and denominators across the dataset and reduce once at the end, matching the way mIoU is reported and remaining stable for rare classes. Boundary-IoU uses a band of width $0.02d$ around each class boundary and is independent of $\tau$ by construction, which the constant B-IoU column
confirms. Predictions are single-scale, taken from the same checkpoints as Table~\ref{tab:seg}. All models are scored on the identical set of 150 classes observed by every model; restricting to this common set is necessary because a class predicted by one model but absent from the ground truth contributes zero to that model's mean, which would otherwise penalize models that predict a wider range of classes.

\textbf{Per-class boundary analysis}.  Aggregate boundary metrics do not indicate where the improvement arises, so we decompose the density effect per class. Per-class estimates are strongly frequency-dependent in variance: among the $43$ classes appearing in fewer than $25$ of the $2{,}000$ validation images the change has standard deviation $7.15$ and only half improve, whereas among the $45$ classes appearing in at least $100$ images the standard deviation falls to $1.83$ and $41$ improve. The mean gain is the same in both the mid- and high-support bins ($+2.29$ and $+2.30$), and gain is uncorrelated with frequency ($r{=}0.05$), so the rare classes contribute noise rather than signal and we restrict \cref{tab:perclass} to well-supported classes. There the pattern is systematic: the largest gains fall on small objects with fine contours---pots, sconces, vases, bottles, streetlights, flowers---while the only losses are large amorphous regions such as mountain, earth, and grass, none exceeding $1.8$ points. Token density improves where a boundary must be placed precisely and is neutral where the boundary is diffuse.

\begin{table}[!htb]
\centering\small\setlength{\tabcolsep}{2pt}
\begin{tabular}{lccc@{\hskip 10pt}lccc}
\toprule
\multicolumn{4}{c}{Largest gains} & \multicolumn{4}{c}{Largest losses} \\
\cmidrule(lr){1-4}\cmidrule(lr){5-8}
Class & \#im & B-IoU & BF & Class & \#im & B-IoU & BF \\
\midrule
pot         & 141 & $+7.21$ & $+9.79$ & mountain & 161 & $-1.78$ & $-1.89$ \\
sconce      & 108 & $+6.64$ & $+6.54$ & earth    & 190 & $-0.80$ & $-0.76$ \\
vase        & 122 & $+5.73$ & $+5.97$ & railing  & 100 & $-0.71$ & $-0.38$ \\
flower      & 138 & $+4.11$ & $+4.78$ & grass    & 227 & $-0.66$ & $-0.29$ \\
bottle      & 114 & $+4.09$ & $+5.98$ &          &     &         &        \\
painting    & 316 & $+4.08$ & $+3.74$ &          &     &         &        \\
streetlight & 240 & $+4.05$ & $+6.57$ &          &     &         &        \\
ashcan      & 100 & $+3.88$ & $+4.55$ &          &     &         &        \\
\bottomrule
\end{tabular}
\caption{\textbf{Per-class boundary change from halving the patch size} at $M{=}32$,
\name-L $p{=}4\!\rightarrow\!p{=}2$ on ADE20K, restricted to the $45$ classes appearing
in at least $100$ validation images. $41/45$ improve (mean $+2.30$ Boundary-IoU,
sd $1.83$).}
\label{tab:perclass}
\end{table}

\begin{table}[t]
\centering
\small
\setlength{\tabcolsep}{2.0pt}
\caption{BF-score under varying boundary tolerance $\tau$ on ADE20K \texttt{val}.
The ranking is unchanged at every tolerance.}
\label{tab:bf-tolerance}
\begin{tabular}{lcccccccc}
\toprule
Backbone & $w$ & $p$ & B-IoU & \multicolumn{5}{c}{BF-score at $\tau/d$} \\
\cmidrule(lr){5-9}
 & & & & 0.002 & 0.004 & 0.008 & 0.016 & 0.032 \\
\midrule
Swin-T          & 8  & 4 & 31.88 & 27.53 & 36.97 & 45.63 & 52.84 & 58.89 \\
SwinV2-T        & 8  & 4 & 32.26 & 27.99 & 37.56 & 46.17 & 53.19 & 59.17 \\
\name-T         & 8  & 4 & 33.11 & 28.81 & 38.61 & 47.36 & 54.48 & 60.47 \\
\midrule
SwinV2-T        & 16 & 4 & 34.14 & 29.59 & 39.50 & 48.22 & 55.13 & 60.74 \\
\name-T         & 16 & 4 & 34.73 & 30.31 & 40.36 & 48.98 & 55.87 & 61.63 \\
\midrule
\name-T         & 32 & 4 & 35.55 & 30.71 & 40.92 & 49.72 & 56.57 & 62.14 \\
\name-T         & 32 & 2 & \textbf{36.42} & \textbf{32.12} & \textbf{42.42}
                          & \textbf{51.06} & \textbf{57.58} & \textbf{62.98} \\
\midrule
\multicolumn{3}{l}{$\Delta$ ($32/2$ vs.\ $32/4$)}
                & $+0.87$ & $+1.41$ & $+1.50$ & $+1.34$ & $+1.01$ & $+0.84$ \\
\bottomrule
\end{tabular}
\end{table}

\subsection{Detection by object size} \label{ssec:detection_by_size}
Table~\ref{tab:detection_size} breaks AP down by object scale, and the two design axes separate cleanly. Widening the window improves every scale, with a mild tilt toward
small objects: over Swin-T, \name-T at $M{=}32$ gains $+19.6\%$ on $\text{AP}_S$ against $+9.3\%$ on $\text{AP}_M$ and $+6.8\%$ on $\text{AP}_L$ for boxes, and $+21.0\%$ /
$+8.7\%$ / $+5.7\%$ respectively for masks. Halving the patch size behaves differently. At fixed $M{=}32$, $p{=}2$ improves $\text{AP}_S$ by $+34.8\%$ for boxes ($20.1 \rightarrow 27.1$) and $+32.6\%$ for masks ($14.4 \rightarrow 19.1$), improves $\text{AP}_M$ by only $+6.9\%$ and $+8.8\%$, and leaves $\text{AP}_L$ essentially untouched ($59.7$ in both cases for boxes, $59.7 \rightarrow 60.2$ for masks). Context and resolution therefore act on different parts of the size distribution: a larger window helps everywhere, consistent with supplying additional surrounding evidence, whereas a finer stem helps only where an object spans too few tokens to be resolved and contributes nothing once it is large enough.

\begin{table*}[!htb]
\centering
\small
\begin{tabular}{lccccccccc}
\toprule
& & \multicolumn{4}{c}{Box} & \multicolumn{4}{c}{Mask} \\
\cmidrule(lr){3-6}\cmidrule(lr){7-10}
Backbone & M/p/Res & AP & AP$_S$ & AP$_M$ & AP$_L$ & AP & AP$_S$ & AP$_M$ & AP$_L$ \\
\midrule
Swin-T      & 8/4/256  & 37.1 & 16.8 & 40.9 & 55.9 & 34.2 & 11.9 & 36.7 & 56.5 \\
SwinV2-T    & 8/4/256  & 37.3 & 17.1 & 41.3 & 55.5 & 34.4 & 12.1 & 37.0 & 56.5 \\
\name-T     & 8/4/256  & \bf 38.5 & \bf 18.1 & \bf 42.9 & \bf 57.2 & \bf 35.3 & \bf 12.8 & \bf 38.2 & \bf 57.2 \\
\midrule
SwinV2-T    & 16/4/256 & \bf 39.0 & 18.7 & \bf 43.3 & 57.5 & 35.7 & 13.2 & 38.6 & \bf 58.7 \\
\name-T     & 16/4/256 & \bf 39.0 & \bf 19.3 & \bf 43.3 & \bf 57.7 & \bf 36.0 & \bf 13.5 & \bf 38.9 & 58.5 \\
\name-T     & 32/4/256 & \bf 40.3 & \bf 20.1 & \bf 44.7 & \bf 59.7 & \bf 36.7 & \bf 14.4 & \bf 39.9 & \bf 59.7 \\
\midrule
\midrule
\name-T     & 32/2/256 & \bf 44.1 & \bf 27.1 & \bf 47.8 & \bf 59.7 & \bf 39.9 & \bf 19.1 & \bf 43.4 & \bf 60.2 \\
\bottomrule
\end{tabular}
\caption{\textbf{COCO detection with size-stratified AP.} Setup as in Tab.~\ref{tab:detection}. Larger windows improve all object scales; the finer patch improves small objects almost exclusively.}
\label{tab:detection_size}
\end{table*}

\begin{table}[]
    \centering
    \small
    \setlength{\tabcolsep}{2pt}
  \begin{tabular}{lcccccc}
\toprule
Route & Patch & Res & GFLOPs & Mem (GB) & img/s$\uparrow$ & IN-1K$\uparrow$ \\
\midrule
finer patch & 2 & 256 & 37.4 & 24.4 & 287 & \textbf{84.1} \\
higher res  & 4 & 512 & 37.4 & 23.8 & 284 & 84.0 \\
\bottomrule
\end{tabular}
\caption{\textbf{Two routes to identical token density.} At matched FLOPs (${\sim}37$G) and parameters (${\approx}28$M) on a $128\!\times\!128$ grid (H200, batch $64$, window $32$), both routes yield equivalent cost and accuracy.}
\label{tab:density-route}
\end{table}

\subsection{Out-of-distribution generalization testing for Large Models.}\label{ssec:out_distribution_large} We evaluate the three Large models without fine-tuning on ImageNet-V2, -A, -R and -Sketch (\cref{tab:ood_large}). Between the two baselines, robustness tracks clean accuracy exactly: SwinV2-L leads Swin-L by $0.42$ points on IN-1K and on every shifted benchmark, by a margin that grows with the severity of the shift ($+0.67$ on V2, $+1.24$ on Sketch, $+1.26$ on -R, $+3.60$ on -A). \name-L breaks that relationship. Its clean accuracy is within $0.02$ of SwinV2-L ($86.91$ vs.\ $86.93$), yet it leads by $0.54$ on ImageNet-V2, $1.10$ on -R and $3.32$ on -A, trailing only on Sketch ($-0.59$). Expressed as degradation from clean accuracy, \name-L loses $8.89$ points on V2 and $17.20$ on -A, against $9.45$ and $20.54$ for SwinV2-L. The advantage is not uniform: SwinV2-L retains a lead on ImageNet-Sketch ($50.48$ vs.\ $49.89$) and on corruption robustness ($37.70$ vs.\ $38.37$ mCE), where \name-L's losses fall mainly in the blur family. Because the two models are separated by hundredths of a point on IN-1K, neither direction can be attributed to fitted accuracy; we report the comparison as an empirical observation without isolating a responsible component.

\begin{table}[!htb]
\centering
\small
\setlength{\tabcolsep}{2pt}
\begin{tabular}{lcccccc}
\toprule
& \multicolumn{5}{c}{Top-1 $\uparrow$} & mCE $\downarrow$ \\
\cmidrule(lr){2-6}\cmidrule(lr){7-7}
Backbone & IN-1K & IN-V2 & IN-A & IN-R & Sketch & IN-C \\
\midrule
Swin-L      & 86.51 & 76.81 & 62.79 & 65.02 & 49.24 & 39.74 \\
SwinV2-L    & \bf 86.93 & 77.48 & 66.39 & 66.28 & \bf 50.48 & \bf 37.70 \\
\name-L     & 86.91 & \bf 78.02 & \bf 69.71 & \bf 67.38 & 49.89 & 38.37 \\
\bottomrule
\end{tabular}
\caption{\textbf{Out-of-distribution generalization at Large scale.} All models are
evaluated without fine-tuning from their IN-22K$\rightarrow$IN-1K checkpoints, each at
its own pretraining window. ImageNet-A and -R are scored over their $200$-class
subsets; ImageNet-V2 uses the matched-frequency split; mCE is AlexNet-normalized over
$15$ corruptions at $5$ severities. At essentially identical clean accuracy
($86.91$ vs.\ $86.93$), \name-L leads on ImageNet-V2, -A and -R.}
\label{tab:ood_large}
\end{table}

\begin{figure}
    \centering
    \includegraphics[width=1.0\linewidth]{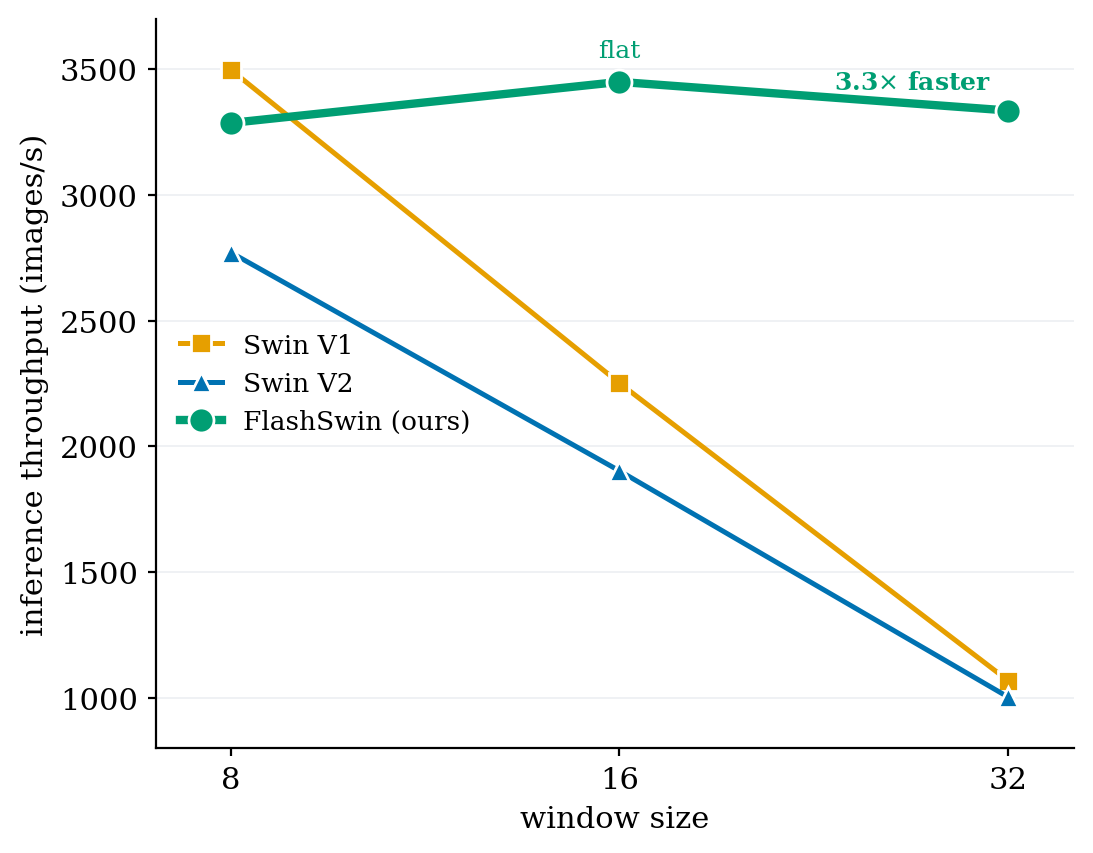}
    \caption{\textbf{Inference throughput versus window size} at patch $4$, $256^2$ input, batch $128$, one H200. \name{}-T is flat across the sweep ($3286\!\rightarrow\!3336$ images/s) while Swin-T falls $3.3\times$ ($3498\!\rightarrow\!1068$) and SwinV2-T $2.8\times$ ($2771\!\rightarrow\!1002$). At $M{=}8$ the fused kernel is slightly slower than the bias-table path, since a small window does not amortise its overhead; \name{} overtakes both baselines from $M{=}16$ onward and is $3.1\times$ faster than Swin-T at $M{=}32$.}
    \label{fig:throughput}
\end{figure}

\begin{figure*}[!thb]
  \centering
    \includegraphics[width=1.0\linewidth]{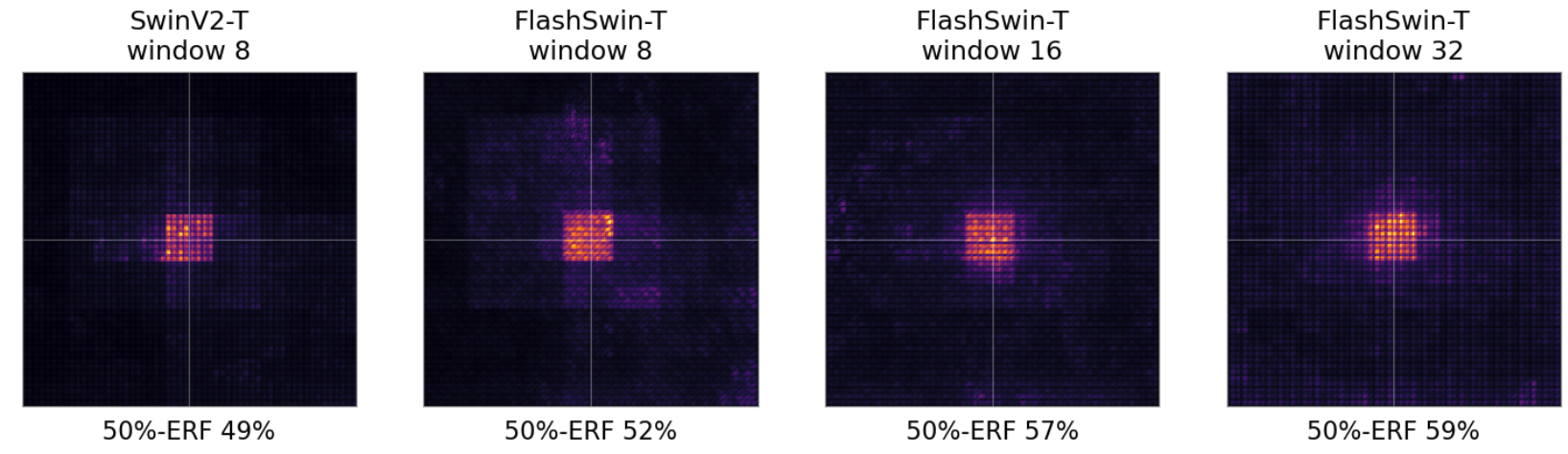}
    \caption{\textbf{Effective receptive field (ERF)} of the final feature at the center location, accumulated over $200$ ImageNet-val images (log scale). \name-T's ERF expands smoothly and coherently as the window grows ($50\%$-ERF radius $50\!\to\!61\!\to\!66\%$ of the input side for $M\!=\!8,16,32$), whereas SwinV2-T ($M\!=\!8$) is markedly more local. Because memory is independent of window size, \name{} can realize the large windows that produce these wide receptive fields.}
\label{fig:erf}
\end{figure*}

\textbf{Receptive field.} \Cref{fig:erf} visualizes the effective receptive field (ERF)~\cite{luo2016understanding}, measuring input-gradient mass of the final center feature across $200$ ImageNet-val images. \name-T's ERF grows monotonically with window size—its $50\%$-mass radius expands from $50\%$ to $66\%$ of the input as $M$ increases from $8$ to $32$—and remains a coherent region without fragmenting. At $M\!=\!8$, \name{} already attains a wider, less strictly local field than SwinV2-T. This directly demonstrates the core mechanism: memory-efficient attention makes large windows affordable, translating into demonstrably larger receptive fields.

\begin{figure*}[!htb]
    \centering
    \includegraphics[width=1.0\linewidth]{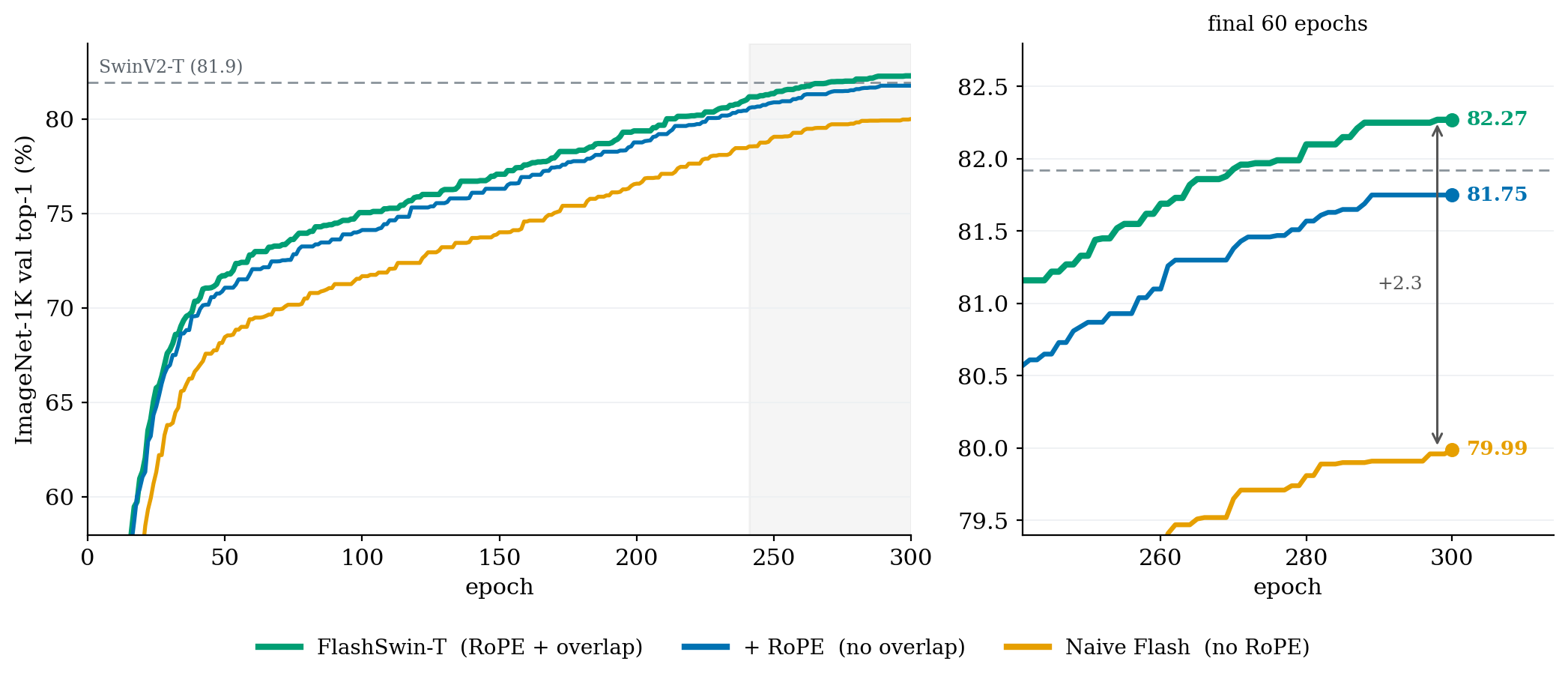}
    \caption{\textbf{Component-ablation training curves} (ImageNet-1K val top-1, w8/p4/256, ${\sim}5.9$ GFLOPs). Naive memory-efficient attention (orange) plateaus below SwinV2-T (dashed, $81.9$); the learnable RoPE (blue) recovers parity and the overlapping stem (green) surpasses it ($+2.3$ total). Right: zoom on the final $60$ epochs with end values.}
\label{fig:ablation-curves}
\end{figure*}

\subsection{What the ROPE learnable frequencies learn.}
\begin{figure}[!htb]
  \centering
    \includegraphics[width=1.0\linewidth]{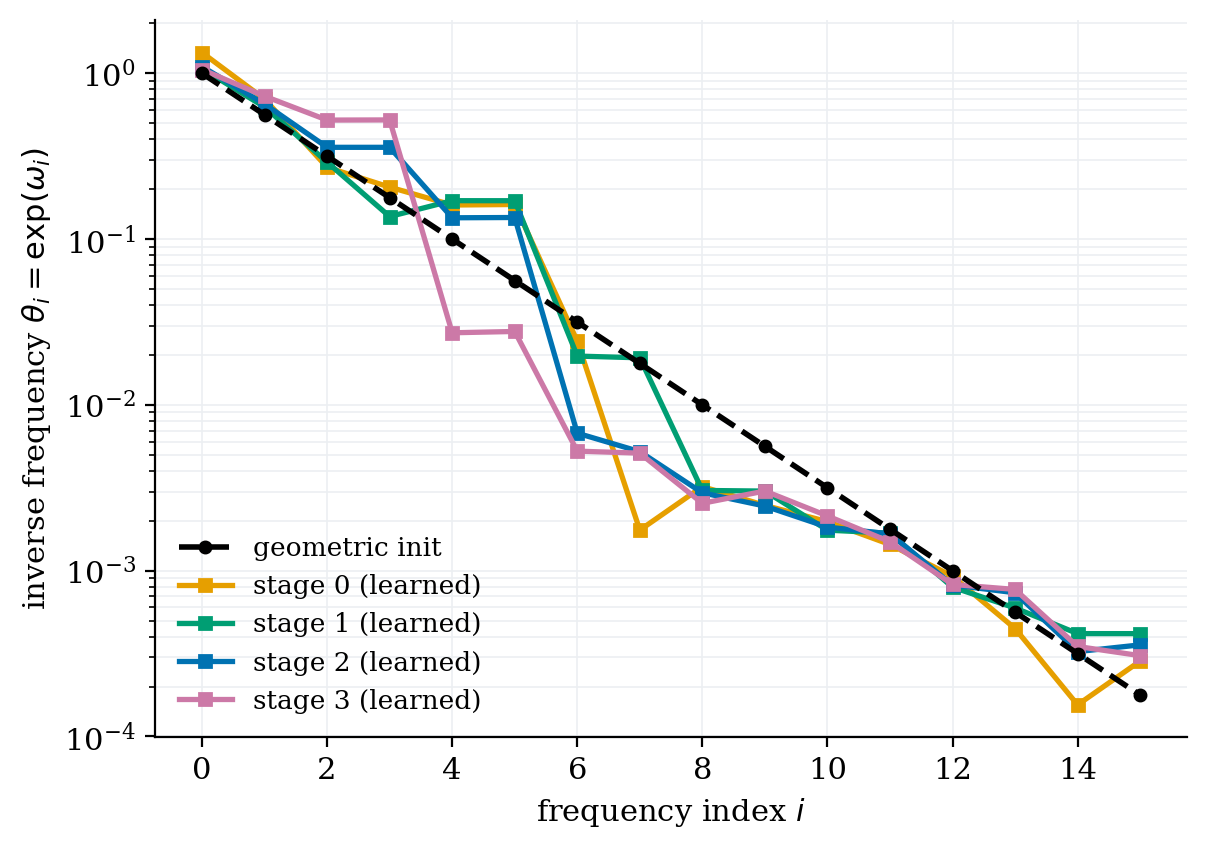}
   \caption{\textbf{Learned RoPE frequencies depart from the geometric initialization.} Inverse frequencies $\theta_i=\exp(\omega_i)$ per stage (averaged over the blocks in each stage) for \name-T, versus the standard geometric RoPE schedule used at initialization (dashed). The learned frequencies deviate markedly---$84\%$ move by more than $10\%$---and each stage adopts a distinct profile, particularly in the mid-band indices, indicating that the optimal positional spectrum is stage-dependent rather than fixed.}
\label{fig:rope-freq}
    
\end{figure}

Because the rotary frequencies are optimized rather than fixed, we can inspect what they converge to. Figure \ref{fig:rope-freq} plots the learned per-stage inverse-frequency spectrum against the geometric initialization. The frequencies do not remain at their initialized values: $84\%$ move by more than $10\%$, and---rather than a single global adjustment---each stage settles on a distinct profile, with the largest departures in the mid-frequency band. This indicates that the positional spectrum best suited to windowed attention is stage-dependent, which a fixed schedule cannot capture, and provides direct empirical support for making the frequencies learnable.

